\documentclass[11pt]{article}
\usepackage{acl2010}
\usepackage{times}
\usepackage{url}
\usepackage{latexsym}

\title{USFD2: Annotating Temporal Expresions and TLINKs for TempEval-2}

\author{Leon Derczynski\\
  Dept of Computer Science\\ 
  University of Sheffield\\ 
  Regent Court\\
  211 Portobello\\ 
  Sheffield S1 4DP, UK\\ 
  {\tt leon@dcs.shef.ac.uk} \And
  Robert Gaizauskas\\
  Dept of Computer Science\\
  University of Sheffield\\
  Regent Court\\
  211 Portobello\\
  Sheffield S1 4DP, UK\\
  {\tt robertg@dcs.shef.ac.uk}
}

\date{}

\begin{document}
\maketitle
\begin{abstract}
We describe the University of Sheffield system used in the TempEval-2 challenge, USFD2. The challenge requires the automatic identification of temporal entities and relations in text. \\USFD2 identifies and anchors temporal expressions, and also attempts two of the four temporal relation assignment tasks. A rule-based system picks out and anchors temporal expressions, and a maximum entropy classifier assigns temporal link labels, based on features that include descriptions of associated temporal signal words. USFD2 identified temporal expressions successfully, and correctly classified their type in 90\% of cases. Determining the relation between an event and time expression in the same sentence was performed at 63\% accuracy, the second highest score in this part of the challenge.
\end{abstract}

\section{Introduction}
The TempEval-2~\cite{pustejovsky2009semeval} challenge proposes six tasks. Our system tackles three of these: task A -- identifying time expressions, assigning {\tt TIMEX3} attribute values, and anchoring them; task C -- determining the temporal relation between an event and time in the same sentence; and task E -- determining the temporal relation between two main events in consecutive sentences. For our participation in the task, we decided to employ both rule- and ML-classifier-based approaches. Temporal expressions are dealt with by sets of rules and regular expressions, and relation labelling performed by NLTK's\footnote{See http://www.nltk.org/ .} maximum entropy classifier with rule-based processing applied during feature generation. The features (described in full in Section~\ref{description}) included attributes from the TempEval-2 training data annotation, augmented by features that can be directly derived from the annotated texts. There are two main aims of this work: (1) to create a rule-based temporal expression annotator that includes knowledge from work published since GUTime~\cite{mani2000robust} and measure its performance, and (2) to measure the performance of a classifier that includes features based on temporal signals.

Our entry to the challenge, USFD2, is a successor to USFD~\cite{hepple2007usfd}. In the rest of this paper, we will describe how USFD2 is constructed (Section~\ref{description}), and then go on to discuss its overall performance and the impact of some internal parameters on specific TempEval tasks. Regarding classifiers, we found that despite using identical feature sets across relation classification tasks, performance varied significantly. We also found that USFD2 performance trends with TempEval-2 did not match those seen when classifiers were trained on other data while performing similar tasks. The paper closes with comments about future work.

\section{System Description}
\label{description}
The TempEval-2 training and test sets are partitioned into data for entity recognition and description, and data for temporal relation classification. We will first discuss our approach for temporal expression recognition, description and anchoring, and then discuss our approach to two of the relation labelling tasks.

\subsection{Identifying, describing and anchoring temporal expressions}
\label{timex}
Task A of TempEval-2 requires the identification of temporal expressions (or {\bf timex}es) by defining a start and end boundary for each expression, and assigning an ID to it. After this, systems should attempt to describe the temporal expression, determining its type and value (described below).

Our timex recogniser works by building a set of n-grams from the data to be annotated ($1 \leq n \leq 5$), and comparing each n-gram against a hand-crafted set of regular expressions. This approach has been shown to achieve high precision, with recall increasing in proportion to ruleset size~\cite{han2006language,mani2000robust,ahn2005towards}. The recogniser chooses the largest possible sequence of words that could be a single temporal expression, discarding any sub-parts that independently match any of our set of regular expressions. The result is a set of boundary-pairs that describe temporal expression locations within documents. This part of the system achieved 0.84 precision and 0.79 recall, for a balanced f1-measure of 0.82.

The next part of the task is to assign a type to each temporal expression. These can be one of {\tt TIME}, {\tt DATE}, {\tt DURATION}, or {\tt SET}. USFD2 only distinguishes between {\tt DATE} and {\tt DURATION} timexes. If the words \emph{for} or \emph{during} occur in the three words before the timex, the timex ends with an {\emph s} (such as in \emph{seven years}), or the timex is a bi-gram whose first token is {\emph a} (e.g. in \emph{a month}), then the timex is deemed to be of type {\tt DURATION}; otherwise it is a {\tt DATE}. These three rules for determining type were created based on observation of output over the test data, and are correct 90\% of the time with the evaluation data.

The final part of task A is to provide a value for the timex. As we only annotate {\tt DATE}s and {\tt DURATION}s, these will be either a fixed calendrical reference in the format YYYY-MM-DD, or a duration in according to the TIMEX2 standard~\cite{ferro2005tides}. Timex strings of \emph{today} or \emph{now} were assigned the special value {\tt PRESENT\_REF}, which assumes that \emph{today} is being used in a literal and not figurative manner, an assumption which holds around 90\% of the time in newswire text~\cite{ahn2005towards} such as that provided for TempEval-2. In an effort to calculate a temporal distance from the document creation time (DCT), USFD2 then checks to see if numeric words (e.g. \emph{one}, \emph{seven hundred}) are in the timex, as well as words like \emph{last} or \emph{next} which determine temporal offset direction. This distance figure supplies either the second parameter to a {\tt DURATION} value, or helps calculate DCT offset. Strings that describe an imprecise amount, such as \emph{few}, are represented in duration values with an {\tt X}, as per the TIMEX2 standard. We next search the timex for temporal unit strings (e.g. \emph{quarter}, \emph{day}). This helps build either a duration length or an offset. If we are anchoring a date, the offset is applied to DCT, and date granularity adjusted according to the coarsest temporal primitive present -- for example, if DCT is 1997-06-12 and our timex is \emph{six months ago}, a value of 1997-01 is assigned, as it is unlikely that the temporal expression refers to the day precisely six months ago, unless followed by the word \emph{today}.

Where weekday names are found, we used Baldwin's 7-day window~\cite{baldwin2002learning} to anchor these to a calendrical timeline. This technique has been found to be accurate over 94\% of the time with newswire text~\cite{mazur2008s}. Where dates are found that do not specify a year or a clear temporal direction marker (e.g., \emph{April 17} vs. \emph{last July}), our algorithm counts the number of days between DCT and the next occurrence of that date. If this is over a limit $f$, then the date is assumed to be last year. This is a very general rule and does not take into account the tendency of very-precisely-described dates to be closer to DCT, and far off dates to be loosely specified. An $f$ of 14 days gives the highest performance based on the TempEval-2 training data.

Anchoring dates / specifying duration lengths was the most complex part of task A and our na\"{\i}ve rule set was correct only 17\% of the time.

\subsection{Labelling temporal relations}
\label{tlink} 

\begin{table}
\begin{center}
\caption{Features used by USFD2 to train a temporal relation classifier.}
\label{tab:features}
\small
\begin{tabular}{| l | r |}
\hline
\textbf{Feature} & \textbf{Type} \\
\hline
\emph{For events} & \\
Tense & String \\
Aspect & String \\
Polarity & pos or neg \\
Modality & String \\
\hline
\emph{For timexes} & \\
Type & Timex type \\
Value & String \\
\hline
\emph{Describing signals} & \\
Signal text & String \\
Signal hint & Relation type \\
Arg 1 before signal? & Boolean \\
Signal before Arg 2? & Boolean \\
\hline
\emph{For every relation} & \\
Arguments are same tense & Boolean \\
Arguments are same aspect & Boolean \\
Arg 1 before Arg 2? & Boolean \\
\hline
\emph{For every interval} & \\
Token number in sentence / 5 & Integer \\
Text annotated & String \\
Interval type & event or timex \\
\hline
\end{tabular}
\end{center}
\end{table}
\normalsize

Our approach for labelling temporal relations (or \textbf{TLINK}s) is based on NLTK's maximum entropy classifier, using the feature sets initially proposed in~\newcite{mani2006machine}. Features that describe temporal signals have been shown to give a 30\% performance boost in TLINKs that employ a signal~\cite{derczynski2010signals}. Thus, the features in~\newcite{mani2006machine} are augmented with those used to describe signals detailed in~\newcite{derczynski2010signals}, with some slight changes. Firstly, as there are no specific TLINK/signal associations in the TempEval-2 data (unlike TimeBank~\cite{pustejovsky2003timebank}), USFD2 needs to perform signal identification and then associate signals with a temporal relation between two events or timexes. Secondly, a look-up list is used to provide TLINK label hints based on a signal word. A list of features employed by USFD2 is in Table~\ref{tab:features}.

We used a simplified version of the approach in~\newcite{cheng2007naist} to identify signal words. This involved the creation of a list of signal phrases that occur in TimeBank with a frequency of 2 or more, and associating a signal from this list with a temporal entity if it is in the same sentence and clause. The textually nearest signal is chosen in the case of conflict.

\small
\begin{table}
\begin{center}
\caption{A sample of signals and the TempEval-2 temporal relation they suggest.}
\label{tab:signalHints}
\begin{tabular}{| l | c |}
\hline
 \textbf{Signal phrase} & \textbf{Suggested relation} \\
\hline
previous & \textsc{after} \\
ahead of & \textsc{before} \\
so far & \textsc{overlap} \\
thereafter & \textsc{before} \\
in anticipation of & \textsc{before} \\
follows & \textsc{after} \\
since then & \textsc{before} \\
soon after & \textsc{after} \\
as of & \textsc{overlap-or-after} \\
throughout & \textsc{overlap} \\
\hline
\end{tabular}
\end{center}
\end{table}
\normalsize

As this list of signal phrases only contained 42 entries, we also decided to define a ``most-likely" temporal relation for each signal. This was done by imagining a short sentence of the form \emph{event1 -- signal -- event2}, and describing the type of relation between event 1 and event 2. An excerpt from these entries is shown in Table~\ref{tab:signalHints}. The hint from this table was included as a feature. Determining whether or not to invert the suggested relation type based on word order was left to the classifier, which is already provided with word order features. It would be possible to build these suggestions from data such as TimeBank, but a number of problems stand in the way; the TimeML and TempEval-2 relation types are not identical, word order often affects the actual relationship type suggested by a signal (e.g. compare \emph{He ran home before he showered} and \emph{Before he ran home, he showered}), and noise in mined data is a problem with the low corpus occurrence frequency of most signals.

This approach was used for both the intra-sentence timex/event TLINK labelling task and also the task of labelling relations between main events in adjacent sentences.

\section{Discussion}
\label{discussion}
USFD2's rule-based element for timex identification and description performs well, even achieving above-average recall despite a much smaller rule set than comparable and more complex systems. However, the temporal anchoring component performs less strongly. The ``all-or-nothing" metric employed for evaluating the annotation of timex values gives non-strict matches a zero score (e.g. if the expected answer is 1990-05-14, no reward is given for 1990-05) even if values are close, which many were.

In previous approaches that used a maximum entropy classifier and comparable feature set~\cite{mani2006machine,derczynski2010signals}, the accuracy of event-event relation classification was higher than that of event-timex classification. Contrary to this, USFD2's event-event classification of relations between main events of successive sentences (Task E) was less accurate than the classification of event-timex relations between events and timexes in the same sentence (Task C). Accuracy in Task C was good (63\%), despite the lack of explicit signal/TLINK associations and the absence of a sophisticated signal recognition and association mechanism. This is higher than USFD2's accuracy in Task E (45\%) though the latter is a harder task, as most TempEval-2 systems performed significantly worse at this task than event/timex relation classification.

Signal information was not relied on by many TempEval 2007 systems (\newcite{min2007lcc} discusses signals to some extent but the system described only includes a single feature -- the signal text), and certainly no processing of this data was performed for that challenge. USFD2 begins to leverage this information, and gives very competitive performance at event/timex classification. In this case, the signals provided an increase from 61.5\% to 63.1\% predictive accuracy in task C. The small size of the improvement might be due to the crude and unevaluated signal identification and association system that we implemented.

The performance of classifier based approaches to temporal link labelling seems to be levelling off -- the 60\%-70\% relation labelling accuracy of work such as~\newcite{mani2006machine} has not been greatly exceeded. This performance level is still the peak of the current generation of systems. Recent improvements, while employing novel approaches to the task that rely on constraints between temporal link types or on complex linguistic information beyond that describable by TimeML attributes, still yield marginal improvements (e.g. ~\newcite{1687936}). It seems that to break through this performance ``wall", we need to continue to innovate with and discuss temporal relation labelling, using information and knowledge from many sources to build practical high-performance systems.

\section{Conclusion}
\label{conclusion}

In this paper, we have presented USFD2, a novel system that annotates temporal expressions and temporal links in text. The system relies on new hand-crafted rules, existing rule sets, machine learning and temporal signal information to make its decisions. Although some of the TempEval-2 tasks are difficult, USFD2 manages to create good and useful annotations of temporal information. USFD2 is available via Google Code\footnote{See http://code.google.com/p/usfd2/ .}.

\section*{Acknowledgments}
Both authors are grateful for the efforts of the TempEval-2 team and appreciate their hard work. The first author would like to acknowledge the UK Engineering and Physical Science Research Council for support in the form of a doctoral studentship.

\bibliographystyle{acl}
\bibliography{usfd2}

\begin{thebibliography}{}

\bibitem[\protect\citename{Ahn \bgroup et al.\egroup }2005]{ahn2005towards}
D.~Ahn, S.F. Adafre, and MD~Rijke.
\newblock 2005.
\newblock {Towards task-based temporal extraction and recognition}.
\newblock In {\em Dagstuhl Seminar Proceedings}, volume 5151.

\bibitem[\protect\citename{Baldwin}2002]{baldwin2002learning}
J.A. Baldwin.
\newblock 2002.
\newblock {\em {Learning temporal annotation of French news}}.
\newblock {Ph.D.} thesis, Georgetown University.

\bibitem[\protect\citename{Cheng \bgroup et al.\egroup }2007]{cheng2007naist}
Y.~Cheng, M.~Asahara, and Y.~Matsumoto.
\newblock 2007.
\newblock {Temporal relation identification using dependency parsed tree}.
\newblock In {\em Proceedings of the 4th International Workshop on Semantic
  Evaluations}, pages 245--248.

\bibitem[\protect\citename{Derczynski and
  Gaizauskas}2010]{derczynski2010signals}
L.~Derczynski and R.~Gaizauskas.
\newblock 2010.
\newblock Using signals to improve automatic classification of temporal
  relations.
\newblock In {\em Proceedings of the ESSLLI StuS}.
\newblock Submitted.

\bibitem[\protect\citename{Ferro \bgroup et al.\egroup }2005]{ferro2005tides}
L.~Ferro, L.~Gerber, I.~Mani, B.~Sundheim, and G.~Wilson.
\newblock 2005.
\newblock {TIDES 2005 standard for the annotation of temporal expressions}.
\newblock Technical report, MITRE.

\bibitem[\protect\citename{Han \bgroup et al.\egroup }2006]{han2006language}
B.~Han, D.~Gates, and L.~Levin.
\newblock 2006.
\newblock {From language to time: A temporal expression anchorer}.
\newblock In {\em Temporal Representation and Reasoning (TIME)}, pages
  196--203.

\bibitem[\protect\citename{Hepple \bgroup et al.\egroup }2007]{hepple2007usfd}
M.~Hepple, A.~Setzer, and R.~Gaizauskas.
\newblock 2007.
\newblock {USFD: preliminary exploration of features and classifiers for the
  TempEval-2007 tasks}.
\newblock In {\em Proceedings of SemEval-2007}, pages 438--441.

\bibitem[\protect\citename{Mani and Wilson}2000]{mani2000robust}
I.~Mani and G.~Wilson.
\newblock 2000.
\newblock {Robust temporal processing of news}.
\newblock In {\em Proceedings of the 38th Annual Meeting on ACL}, pages 69--76.
  ACL.

\bibitem[\protect\citename{Mani \bgroup et al.\egroup }2006]{mani2006machine}
I.~Mani, M.~Verhagen, B.~Wellner, C.M. Lee, and J.~Pustejovsky.
\newblock 2006.
\newblock {Machine learning of temporal relations}.
\newblock In {\em Proceedings of the 21st International Conference on
  Computational Linguistics}, page 760. ACL.

\bibitem[\protect\citename{Mazur and Dale}2008]{mazur2008s}
P.~Mazur and R.~Dale.
\newblock 2008.
\newblock {What’s the date? High accuracy interpretation of weekday}.
\newblock In {\em 22nd International Conference on Computational Linguistics
  (Coling 2008), Manchester, UK}, pages 553--560.

\bibitem[\protect\citename{Min \bgroup et al.\egroup }2007]{min2007lcc}
C.~Min, M.~Srikanth, and A.~Fowler.
\newblock 2007.
\newblock {LCC-TE: a hybrid approach to temporal relation identification in
  news text}.
\newblock In {\em Proceedings of the 4th International Workshop on Semantic
  Evaluations}, pages 219--222.

\bibitem[\protect\citename{Pustejovsky and
  Verhagen}2009]{pustejovsky2009semeval}
J.~Pustejovsky and M.~Verhagen.
\newblock 2009.
\newblock {SemEval-2010 task 13: evaluating events, time expressions, and
  temporal relations (TempEval-2)}.
\newblock In {\em Proceedings of the Workshop on Semantic Evaluations}, pages
  112--116. ACL.

\bibitem[\protect\citename{Pustejovsky \bgroup et al.\egroup
  }2003]{pustejovsky2003timebank}
J.~Pustejovsky, P.~Hanks, R.~Sauri, A.~See, R.~Gaizauskas, A.~Setzer, D.~Radev,
  D.~Day, L.~Ferro, et~al.
\newblock 2003.
\newblock {The Timebank Corpus}.
\newblock In {\em Corpus Linguistics}, volume 2003, page~40.

\bibitem[\protect\citename{Yoshikawa \bgroup et al.\egroup }2009]{1687936}
K.~Yoshikawa, S.~Riedel, M.~Asahara, and Y.~Matsumoto.
\newblock 2009.
\newblock Jointly identifying temporal relations with markov logic.
\newblock In {\em IJCNLP: Proceedings of 47th Annual Meeting of the ACL}, pages
  405--413.

\end{thebibliography}

\end{document}